%% file: approxHW.tex
\documentclass[11pt,a4paper]{article}

\usepackage[bf,sf]{titlesec}
\usepackage[utf8]{inputenc}
\usepackage{verbatim}
\usepackage{graphicx}
\usepackage{hyperref}
\usepackage{url}
\usepackage[numbers,square]{natbib}
\usepackage{amsmath, amssymb}
\usepackage{verbatim}
\usepackage{algorithm}
\usepackage{algpseudocode}
\usepackage[skip=0pt]{caption} 
\usepackage{titlesec}

\usepackage{fullpage}

\usepackage{fancyhdr}

\usepackage{bm}
\usepackage{amsmath}

\usepackage{ifpdf}

\usepackage{tgpagella}
\usepackage[T1]{fontenc}
\usepackage{fancyhdr}   
\usepackage{makeidx}

\newcommand{\reals}{\mathbb{R}}

\ifpdf
\DeclareGraphicsExtensions{.png, .pdf, .jpg, .tif}
\else
\DeclareGraphicsExtensions{.eps, .jpg}
\fi

\input{macros.tex}

\title{{\bf Approximate Hubel-Wiesel Modules and the\\ Data Structures of Neural Computation}}
\author{Joel Z Leibo, $~~~$Julien Cornebise, $~~~$Sergio G\'{o}mez,  $~~~$Demis Hassabis}

\date{Google DeepMind, London UK}

\begin{document}

\maketitle

\begin{abstract}
This paper describes a framework for modeling the interface between perception and memory on the algorithmic level of analysis. It is consistent with phenomena associated with many different brain regions. These include view-dependence (and invariance) effects in visual psychophysics \citep{bulthoff1992psychophysical, Tarr1995} and inferotemporal cortex physiology \citep{Hung2005, freiwald2010functional}, as well as episodic memory recall interference effects associated with the medial temporal lobe \citep{Tulving1985, Andersen2006}. The perspective developed here relies on a novel interpretation of Hubel and Wiesel's conjecture for how receptive fields tuned to complex objects, and invariant to details, could be achieved \citep{Hubel1962}. It complements existing accounts of two-speed learning systems in neocortex and hippocampus (e.g., \cite{McClelland1995, Norman2003}) while significantly expanding their scope to encompass a unified view of the entire pathway from V1 to hippocampus.  
\end{abstract}

\section{Introduction}

An associative memory can be seen as a data structure wherein stored items are mapped to labels. Or equivalently, items are grouped into sets $C_k$, each consisting of all the items sharing the same label. If the items are input feature vectors and the sets correspond to episodic memories then this is a standard view of (part of) the hippocampus. A generalization of an associative memory that is appropriate in any metric space maps items---represented as vectors of distances to stored exemplars---to numbers. The output of this map may be interpreted as the item's degree of membership in a particular $C_k$. If the items are feature vectors, the exemplars are simple cell receptive fields, and the $C_k$ are complex cell receptive fields, then this is a standard view of primary visual cortex.

In this article, we explore the implications of adopting a unified view of hippocampus and the ventral visual pathway based around this data structure analogy. This is a fairly radical idea in light of much of the literature in these areas. The computational and experimental neuroscience communities have historically treated perception and memory as almost entirely separate fields. Episodic memory is critically supported by the hippocampus and the rest of the medial temporal lobe (MTL) \citep{Tulving1985, Andersen2006}, thus models of this type of declarative memory have tended to concentrate on these structures e.g.,  \cite{Norman2003}. Conversely models of the ventral visual processing stream typically span from V1 to inferotemporal cortex (IT) e.g., \cite{Serre2007}, the last unimodal area in the ventral visual pathway. However, IT cortex is in fact only one synapse away from perirhinal cortex (PRC) in the MTL \citep{Suzuki1994a}. Yet despite this close anatomical proximity, researchers focusing on either side of this divide are separated by a wall of terminological, methodological and cultural differences. Here we argue that from a computational perspective this barrier is actually largely artificial and arbitrary, and therefore should be broken down.

A major limitation of nearly all computational models of the MTL to date (e.g., \cite{Norman2003, Kali2004}) is the simplifying assumptions made regarding the inputs to the memory system. Most models use abstract input features (e.g., arbitrary binary patterns) for reasons of convenience rather than attempting to accurately capture particular properties of the perceptual representations in the afferents to the MTL. Because these abstracted representations cannot be computed from real sensory data, most MTL models are limited to highly simplified situations. The natural place to look for models of perception capable of bridging this gap is the literature on computational models of the ventral stream e.g., \cite{Riesenhuber1999, Lee2003, Rolls2012, DiCarlo2012}.

Unified models of neural computation are often motivated by neocortex's profound plasticity in cases of missing e.g., congenital blindness \citep{sadato1996activation, roder2002speech} or rewired inputs \citep{sur1988experimentally}. Under these proposals the differing properties of brain regions arise due to the different inputs they receive, as opposed to their implementing fundamentally different algorithms. Such unified models include convolutional neural networks \citep{LeCun1989} and their relatives \citep{Fukushima1980,Riesenhuber1999} and have influenced the current state of the art in machine learning \citep{Krizhevsky2012, Abdel-Hamid2012}. However, one should be skeptical they can be straightforwardly extended into the MTL. Unlike the six layered neocortex of the ventral stream, hippocampus is made up of a quite different three-layered structure called archicortex.  If there is any point in this extended cortical hierarchy where a different algorithm is most likely to lurk, the transition from entorhinal cortex to hippocampus would be a good bet. The unified view we propose here is mostly on the algorithmic level of analysis \citep{marr2010vision}. As such, we argue that the brain implements at least two different approximations to the same underlying data structure. The specific approximation that is most suitable will turn out to differ between neocortex and hippocampus. 

The rest of this paper proceeds as follows. In section~\ref{section:associative_decision_makers} we motivate a particular biologically plausible hierarchy of modules. Then we consider how its connections could be learned from data. In section \ref{section:HW_modules} we consider an idealized learning process. This ideal itself is not biologically plausible, but the new metaphor for  neural computation introduced in section~\ref{section:data} motivates several approximations which are plausible. These are described in section \ref{sec:approx}. Next, we point out that different approximations will turn out to be suitable for different timescales of learning, i.e., fast in hippocampus and slow in neocortex. The final section describes simulations of one model arising from this perspective.   

\section{A Hierarchy of Associators} \label{section:associative_decision_makers}
It is well-known that the sample-complexity of associative learning can be decreased by incorporating assumptions on the space of hypotheses. One way to do this is to employ preprocessor modules that associate together regions of the state space that ought to be treated the same way. The argument is recursive, from the point of view of the preprocessor module: it too could be learned with lower sample complexity if it had its own preprocessor, and so on. But what is the optimal way to grow out this hierarchy of subordinate associators? How many levels should there be? How many branches per level? Clearly, these answers are determined by the environment itself. Therein lies the fundamental problem: the environment cannot be known in any way but statistically. Since the world is stochastic, the only way to learn what states ought to be associated is to wait until enough information has come in. 
In this context, there are two arguments for why brains need a two-speed memory system, i.e., a hippocampus and a neocortex.
\begin{enumerate}
\item If a hierarchy of modules must all learn and act at once in a non-stationary world, then the higher level module must always operate on a faster timescale than its subordinates. If it did not, the changing semantics of its inputs would quickly render its own associations useless. 
\item If every stationary aspect of the environment is ``filtered out'' by the lower modules, the input passed to the associator at the highest level will only contain the information specific to the current timestep. There is therefore no need for the highest level to accumulate more than one timestep before deciding. Thus there is, in principle, no limit to the speed at which it can decide.  
\end{enumerate} 
In a continuous state-space, an associator clustering its space into $K$ regions maps each state $x \in \mathcal{X} = \reals^d$ to a vector $\mu(x) = (\mu_{1}(x), \hdots, \mu_{K}(x))$ called the \emph{signature} of $x$.  This signature function $\mu:\reals^d \rightarrow \reals^K$  represents a graded membership to each of $K$ classes.  Each class is characterized by a particular set $\mathcal{T}_{k} \subset \mathcal{X}$, called a \emph{template book}.  For a pooling function $P:\reals^{| \mathcal{T}_k |} \rightarrow \reals$, invariant to the order of its arguments, and a similarity function $f:\mathcal{X}\times\mathcal{X} \rightarrow \reals$, the $k$-th element of the signature is:
\begin{equation}
\label{eq:exact_HW-module}
\mu_k(x) =  P (\left\{f(x , t) : t\in \mathcal{T}_k \right\}).
\end{equation}
We typically choose $P$ to be $\max$ or $\Sigma$, but many other functions are also possible (see \citep{PartI2013}). In the case of $P=\max$, the signature can be thought of as the vector of similarities of $x$ to the most similar item in each of $K$ sets of stored templates. If a hard membership is required, then $\argmax_{k} \mu_{k}(x)$ will assign $x$ to one of the $K$ classes. Every level of the hierarchy of modules computes its own signature of the input. The inputs to layer $\ell+1$ are encoded by their signatures at layer $\ell$. 

\section{Hubel-Wiesel Modules} \label{section:HW_modules}

The connectivity required to implement a layer of associators is thought to exist in at least one part of the brain. Take $f(x,t)$ to be the response of a V1 simple cell, and all the $t \in \mathcal{T}_k$ to be different simple cell receptive fields sharing the same preferred orientation but differing in their preferred position. Then, assuming Hubel and Wiesel's conjecture for the connectivity of simple (abbr. S) and complex (abbr. C) cells is at least approximately true, $\mu_k(x)$ is the response of the C-cell that pools that set of S-cells \citep{Hubel1962}.

We call the neural network consisting of one C-cell and all its afferent S-cells a \emph{Hubel-Wiesel module} abbr. \emph{HW-module}. We assume that $\|t\| =1 ~ (\forall t)$. For a nonlinearity $f$, let $S = \{f(x , t) ~ : ~ \forall $t$ \in \cup_k \mathcal{T}_k \}$ be the set of S-cell responses to $x$. The function $f$ could be the sigmoid of the dot product, i.e., $f(x,y) = \sigma(f \cdot t)$ as in classical neural networks. However, in most of the present work we use a normalized dot product $f(x,t) = (x \cdot t) / \|x\|$ so we can use the intuition that it measures the similarity between $x$ and $t$.  An HW-module is a reusable network motif. It appears as an essential architectural element in a wide range of machine learning systems including convolutional neural networks \citep{LeCun1989,lecun1995convolutional}, HMAX \citep{Riesenhuber1999,Serre2007a}, and nearest neighbor search \citep{Fix1951}. One \emph{HW-layer} consists of many HW-modules. An \emph{HW-architecture} consists of many HW-layers stacked in a deep hierarchy.

Notice that this notation treats parametric and nonparametric models the same way. In the former case, the templates $t \in \cup_k \mathcal{T}_k$ would usually be obtained by optimization with respect to a loss function. In the nonparametric case, they can be thought of as training data points (or functions of them). For nonparametric supervised learning, the template book $\mathcal{T}_k$ could be the subset of the training data with label $k$. In the unsupervised parametric case, learning the $\mathcal{T}_k$ corresponds to learning ``feature pooling'' e.g., it could be done by agglomerative clustering \citep{coates2012emergence}. In the unsupervised nonparametric case, any clustering algorithm, or alternatively, temporal-continuity-based associative methods like \citep{Isik2012, liao2014unsupervised} could be used to assign training examples to template books.

\subsection{Invariance and the ventral stream}\label{subsection:invariance}
The different sensory streams have different computational goals. In the case of the ventral stream, the computational goal is to enable visual object recognition---the crux of this problem being that of computing an invariant and discriminative representation for unfamiliar objects. One way to construct such a representation with an HW-architecture is to let the template book $\mathcal{T}_k$ be the orbit of a \textit{base template} $t_k$ under the action of a group $G$. Use $g$ to denote a (unitary) representation of an element of $G$ (by an abuse notation it can indicate the corresponding group element as well). The $k$-th template book is then $\mathcal{T}_k = \{gt_k : g \in G\}$. This can be regarded as the outcome of an idealized temporal association process \citep{Foldiak1991, Isik2012, PartI2013}.

The orbit is invariant with respect to the group action and unique to each object \citep{PartI2013}. For example, the set of images obtained by rotating $x$ is the same set of images obtained by rotating $gx$. Thus the set of scalar products between $x$ and all the $gt_k \in \mathcal{T}_k$ is the same as the set of scalar products between $gx$ and $gt_k \in \mathcal{T}_k$, though their order will be permuted. Since the pooling function $P$ is unaffected by permuting the order of its arguments, the output of an HW-module with a group-generated template book is invariant to the action of $G$ \citep{PartI2013,Liao2013}\footnote{Notice also that an HW-module could pool over a subset of the orbit \citep{PartI2013}. This gives a way to model neurons that respond maximally to their preferred feature at any position within a receptive field having some limited spatial extent. The canonical examples are C-cells in V1, i.e., the same cells that motivated the HW-module notion in the first place. A  classic convolutional neural net, e.g., in the sense of \cite{lecun1995convolutional}, is obtained by choosing $G$ = translations (or a subset). In that case, each $gt_k$ will be a copy of $t_k$ shifted to a different position.}. 

\subsection{Binding and multimodal invariance in the medial temporal lobe} \label{subsection:binding}
In section \ref{section:associative_decision_makers} we argued that an associator's burden of sample complexity could be lessened by employing a preprocessor consisting of a set of subordinate associators. Priors for wiring up the HW-modules comprising each may be chosen so that particularly useful aspects of the environment are highlighted. However, it is possible that the very fact of being a good subsystem for pulling out some environmental property makes a system worse at pulling out others. For example, information about an object's identity and its position may conflict in this way. It seems that a system capable of making arbitrary associations, like the hippocampus, is needed in such cases. We propose, in accord with \citep{eichenbaum1993memory}, that the hippocampus's ability to quickly make arbitrary associations makes it likely to be a key player in the binding together of representations from the ventral and dorsal visual streams. It could play a similar role with respect to cross modal representations \citep{quiroga2009explicit}.

\section{The Data Structures of Neural Computation}\label{section:data}

The usual metaphor for a multistage feedforward neural computation is a chain of representation transformations. Each stage, which might correspond to a brain region, e.g., V1, V2, V4, is regarded as a function taking the neural activity vector representing a stimulus in the language of the previous stage as its input and returning to the next stage a transformation of it \citep{Fukushima1980, riesenhuber2002neural, DiCarlo2012}. This idea of a transformation cascade has been useful as a description of the ventral stream and as a motivator for computer vision algorithms. However, it may not be rich enough to naturally accommodate the range of phenomena one would like to model in the MTL literature. Now we explore an alternative metaphor for multistage feedforward computation. Rather than the central question being: what is the chain of transformations? Our proposal asks: what are the data structures implemented by cortex at each stage? 

An HW-module can be viewed as a data structure. In this view, an HW-module consists of a set of data values $\mathcal{D}$ and associated operations for accessing and manipulating them. For example, one could be the tuple $(\mathcal{D},\texttt{INSERT},\texttt{QUERY})$. \texttt{INSERT} is an abstraction of learning. After obtaining experience with a new stimulus, its representation gets inserted in the correct format to one or more HW-modules.  The inference process at test-time is a cascade of \texttt{QUERY} operations. The input stimulus is used to query the first stage HW-modules which return---rather, pass along to query the next stage---a result that depends on the relationship between the input and the stored data. Other operations such as \texttt{DELETE} are also possible, indeed this extensibility is one of the motivations for adopting the data structure interpretation. However, the scope of the present paper will be limited to the two basic operations. 

For example, the following describes an HW-architecture implementing a $K$-way 1-nearest neighbor classifier.  Each of the $K$ HW-modules stores a set of data $\mathcal{D}$, and comes with two functions \texttt{INSERT} and \texttt{QUERY}. The data held in the $k$-th HW-module $\mathcal{D}_k$ is the template book $\mathcal{T}_k$ consisting of the set of examples with label $k$: $\mathcal{D}_k = \mathcal{T}_k = \{t^1_k, \dots, t_k^n\}$, where the $k$-th template book $\mathcal{T}_k$ is the set of examples with label $k$. For a new $k$-labeled example $t \in \reals^d$, and an input $x \in \reals^d$,
\begin{align}
\texttt{INSERT}(\mathcal{D}_k, t) :  ~ \mathcal{D}_k & \leftarrow \mathcal{D}_k \cup \{t\}  \label{NN_insert}\,.\\
\texttt{QUERY}(\mathcal{D}_k, x)  :  ~   \mu_k(x)  & \leftarrow \max_{t\in \mathcal{D}_k}  \langle x , t \rangle \,.  \label{NN_query}
\end{align}
The predicted category is $\hat{y}(x)  :  ~  \leftarrow \argmax_{k = 1 \hdots K} \mu_k(x)$.

In conjunction with the data structure perspective, nonparametric models like nearest neighbors can capture episodic memories naturally. To remember a specific stimulus, like a phone number, just \texttt{INSERT} it. It is not as clear how to get such behavior with parametric models. Thus, for the remainder of this paper we restrict our discussion to the nonparametric case.

\section{Approximate HW-Modules}\label{sec:approx}

So far we have discussed an idealized case, we may call it an  \textit{exact} HW-architecture. Its \texttt{INSERT} operation  \eqref{NN_insert} is not biologically plausible. We propose instead that each stage of the brain's feedforward hierarchy implements an approximation to it. Different approximation methods may be used in different stages.

\subsection{Two biologically plausible approximations} \label{section:two_approx}
Let $\mathbb{T}_k$ and corresponding to the template book $\mathcal{T}_k$. Each template $t_k \in \mathcal{T}_k$ is a row of $\mathbb{T}_k$. Choose $f$ to be a normalized dot product. If $x$ and $t$ are normalized,  the vector $\vec{S}_k(x)$ computed by the S-cells of the $k$-th HW-module is just $\vec{S}_k (x) = \mathbb{T}_k x$.

The best rank-$r$ approximation of $\mathbb{T}_k$ is its singular value decomposition (SVD) $\hat{\mathbb{T}}_k \approx U \Sigma V^\intercal$,
where $U \in \reals^{\|\mathcal{T}_k| \times r}$, $\Sigma \in \reals^{r \times r}$, and $V \in \reals^{d \times r}$. Let $[\mathbb{T}_k | t]$ indicate the concatenation of $t$ as an extra row of $\mathbb{T}_{k}$. Thus, if each S-cell stores a row of $\mathbb{T}_k V$, the best rank-$r$ approximation of the exact HW-module is \mbox{$(\mathcal{D}_k = \mathbb{T}_k V, \texttt{INSERT}, \texttt{QUERY})$}:
\begin{align}
\texttt{INSERT}(\mathcal{D}_k,t) : ~  \mathcal{D}_k &\leftarrow [\mathbb{T}_k | t] V^\prime ~~~~~~~~~~~~~~ \text{ with } U^\prime \Sigma^\prime V^{{\prime}^\intercal} = [\mathbb{T}_k | t] \\
\texttt{QUERY}(\mathcal{D}_k, x) : ~ \hat{\mu}_k(x) &\leftarrow  \max_{i = 1, \ldots, |\mathcal{T}_{k}|} \left(\mathbb{T}_k V V^\intercal x \right)_{i}
\end{align}
Any online PCA algorithm could be used to update $\mathbb{T}_k V$ as new data is inserted. The most biologically plausible is the learning rule proposed by Oja as an approximation to the normalized Hebbian rule \citep{oja1982simplified}\footnote{Oja's rule converges to a solution network that projects new inputs onto the first eigenvector of the past input's covariance, i.e., onto the first column of $V_r$. In the presence of noise, Oja's rule may also give other eigenvectors. There are also modifications of Oja's basic rule that find as many eigenvectors as desired \citep{sanger1989optimal, oja1992principal}.}. Oja's rule provides a biologically plausible way to implement \texttt{INSERT}. However, it generally takes several epochs of looping through the same data items before it converges. Next we discuss another approximation strategy that---while it does not yield the best rank-$r$ approximation---supports rapid insertion of new data.

Since random projections may preserve dot products (as in the Johnson-Lindenstrauss theorem \citep{Johnson1984}), it is also possible to approximate the S-layer response vector by $\vec{S}(x)  \approx \mathbb{T}_k R R^\intercal x$ where $R$ is a $d \times s $ random matrix satisfying the hypotheses of the Johnson-Lindenstrauss theorem (with $s \ll d$). This approximation will generally be less efficient than the PCA approximation obtained from Oja's rule. However, it has a fast \texttt{INSERT} operation:
\begin{equation}
\texttt{INSERT}(\mathcal{D}_k,t ) : \mathcal{D}_k ~ \leftarrow [\mathbb{T}_k | t] ~ [R | r] ~~~~~~~~~ \text{ where } r = \text{random vector s.t. $[R|r]$ orthogonal}
\end{equation}

Increasing the rank of $RR^\intercal$ with each insertion allows the HW-module to store arbitrary amounts of data without running into the Johnson-Lindenstrauss bound \citep{Johnson1984}. Alternatively, it might only augment $R$ when near the bound.


The dichotomy between these two biologically plausible \texttt{INSERT} operations motivates an interesting conjecture concerning complementary ``fast'' and ``slow'' learning systems in the hippocampus and neocortex \citep{McClelland1995}. The PCA approximation, implemented by Oja's rule, could operate in cortex while a random-projection-based approximation could be the approximation used in the hippocampus. Since the latter requires the creation of new random vectors, this could be why there is neurogenesis in dentate gyrus and why it is not needed in neocortex. 

This conjecture can be seen as a revision of McClelland et al.'s complementary learning systems proposal. However, in our case, the reason for the two learning systems is not to cope with catastrophic interference. Instead, we highlight that cortex, which deals with more constrained tasks, can implement strong priors appropriate for each one. For example, the temporal continuity of object motion can be leveraged toward unsupervised learning of invariant representations \citep{Foldiak1991, wiskott2002slow}. Hippocampus however, must be able to make arbitrary associations. Thus its \texttt{INSERT} operation must work even in cases where a stimulus is only encountered once, and does not necessarily have any similarity to previously stored items. A random projection scheme, able to immediately encode a new item, can do this.

\subsection{Locality sensitive hashing-based approximation}
HW-architectures with certain parameter settings (i.e., filter sizes, pooling domains, etc) are equivalent to convolutional neural networks, and with other parameter settings are equivalent to nearest neighbor search algorithms. This correspondence suggests a powerful approximation strategy. It may not be biologically plausible in its details but it is interesting nonetheless. It shares more in common with the random projection strategy than the PCA, thus if it were used by the brain, hippocampus would be the most likely place.

Assume max-pooling for all the following.  Locality-sensitive hashing (LSH) is a data structure that supports fast queries by solving an approximate nearest neighbor problem. It can be recast as a data structure for fast querying of HW-modules. Thus it can approximate max-pooling convolutional neural networks analogously to the way it approximates nearest neighbor search.

Many different LSH schemes exist. Inspired by the impressive results of \citep{Dean2013}, we chose winner-take-all (WTA) hashing \citep{Yagnik2011} for the implementation we used in our experiments.

\begin{figure}[t]
\begin{minipage}{.33\textwidth}
\centering
\includegraphics[width=1\textwidth]{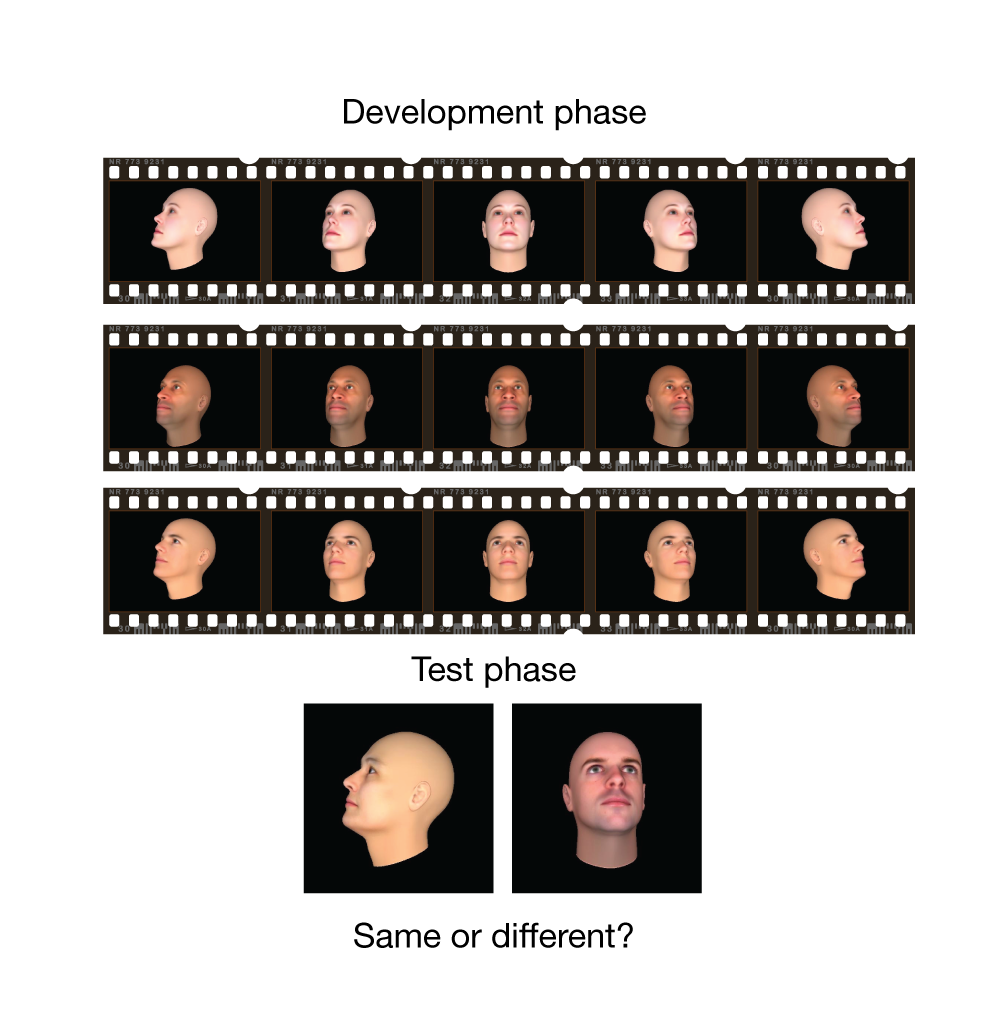}%
\captionof{figure}{Examples of faces used in our experiment, here with uniform background.} 
\label{fig:faces}
\end{minipage}\hfill
\begin{minipage}{.55\textwidth}
\centering
\includegraphics[width=1\textwidth]{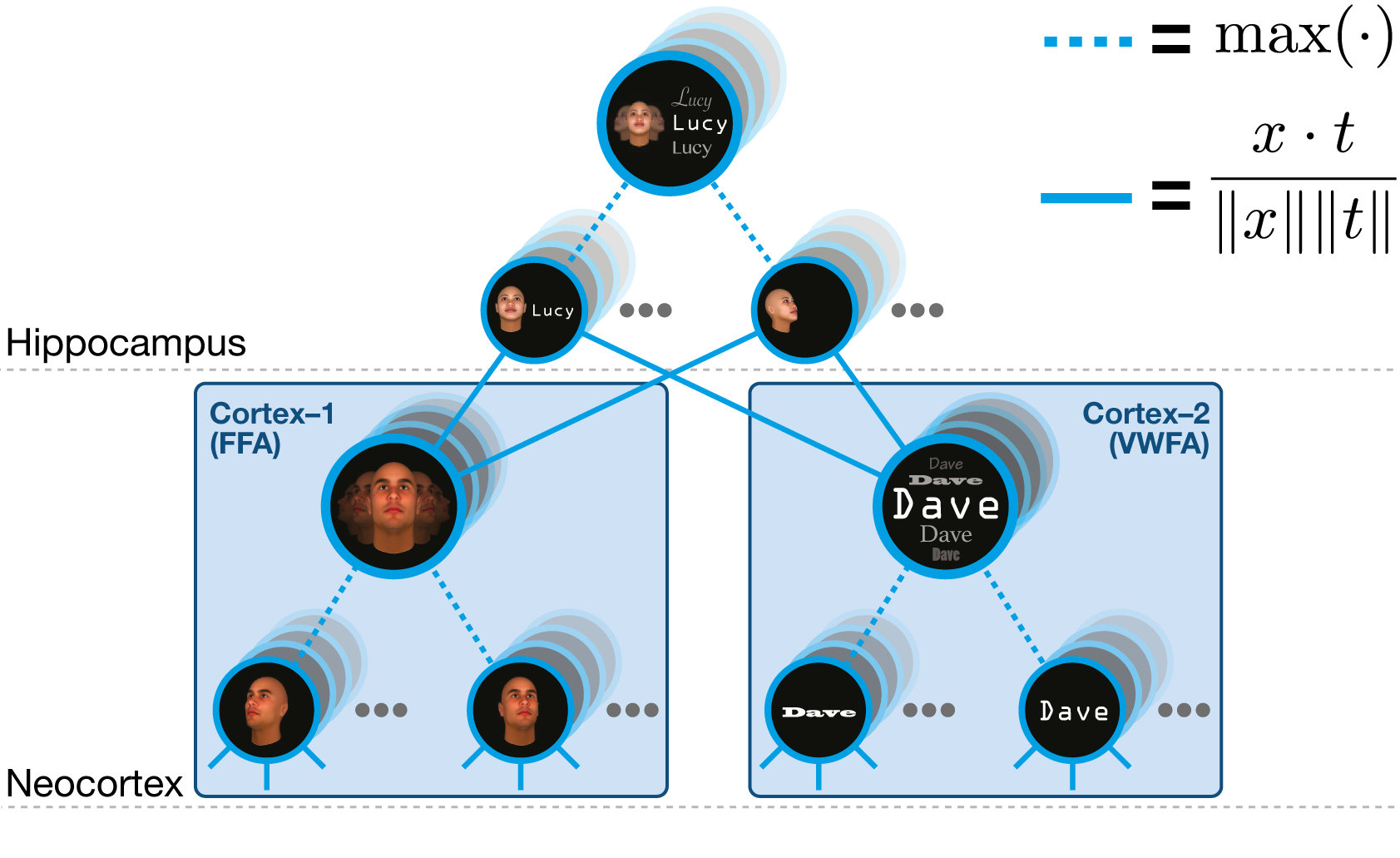}
\captionof{figure}{Two-layered HW-architecture, as described in section~\ref{section:experiments}.}
\label{fig:arch}
\end{minipage}
\end{figure}

\section{Approximation Algorithm and Model Architecture}\label{section:algorithm}
The typical architecture is illustrated in figure \ref{fig:arch} and used in section~\ref{section:experiments}. Its consists of two layers of HW-modules: upper layer models cortex (PCA), lower models hippocampus (WTA). The cortical layers used the PCA approximation and the hippocampal layer used the WTA-based approximation described in Algorithm \ref{alg:insertquery}.

The data stored in one HW-module is now $\mathcal{D}_k = \left(\mathcal{T}_{k}, \mathcal{H}_{k}\right)$ coupling stored templates with their hashes. An HW-layer is a set of $K$ HW-modules. The creation of the candidates set $C_k$ in Line~\ref{lin:candidates} of Algorithm~\ref{alg:insertquery} can be done in several ways, with parallelized explicit comparison of the integer-valued hashes for all templates, or via binning and two-stage indexing as in E2LSH~\citep{Andoni2006} when the saving on the larger number of  templates compensates the runtime overhead.

\begin{algorithm}[htbp]
\caption{Insertion and querying of an approximate HW-layer  \label{alg:insertquery}}
\begin{algorithmic}[1]
\Function{$\texttt{INSERT}$} {$D_k, t$}
\State  $\mathcal{T}_k \leftarrow \mathcal{T}_{k} \cup \{t\}$ 
\State  $\mathcal{H}_{k} \leftarrow \mathcal{H}_{k} \cup \{( h_{1}(t), \ldots, h_{L}(t) )\}$ \Comment{Use $L$ hashes for amplification} 
\EndFunction
\Function{$\texttt{QUERY}$}{$D_{k}, x$}
\State $C_{k} \leftarrow \bigcup_{i=1}^{L} \{t \in \mathcal{T}_{k} : h_{i}(x) = h_{i}(t)\}$ \hspace{.05in} using pre-computed $\mathcal{H}_{k}$ \Comment{See text} \label{lin:candidates}
\State $\mu_{k}(x) \leftarrow P ( \left\{f( x , t ) : t\in \mathcal{C}_k \right\}) $ \Comment{Parallelized by GPU, approximates \eqref{eq:exact_HW-module}}
\State \Return $\mu_{k}(x)$
\EndFunction
\end{algorithmic}
\end{algorithm}

\section{Experiments} \label{section:experiments}

We present two experiments illustrating respectively our Ventral Stream and our MTL models. They share most of their architecture, illustrated in figure~\ref{fig:arch}, using the algorithms of section~\ref{section:algorithm}. Cortex has two subdivisions. Cortex-1, used for both experiments, models a face-selective region like the fusiform face area (FFA). The S-units of cortex-1 are tuned to images of faces at different angles. There is one HW-module for each familiar individual. Cortex-2, used only for the ventral stream experiment, models a word-selective region like the visual word form area (VWFA). Its S-cells are tuned to images of written names. Within one HW-module, all S-cells are tuned to images of names in different fonts and at different retinal locations. S-cells in the hippocampus are tuned to specific associations in the previous layer. Each hippocampal S-cell is connected to \emph{all} C-cells in the cortex: C1 only  for ventral stream experiment, and C1 and C2 for MTL experiment. Thus the templates stored by the S-cells are distributed representations over the entire layer below.
There is one HW-module in the hippocampus for each individual person = \{face, name\} to be remembered. 

\subsection{Ventral Stream experiment: Same-different matching of unfamiliar faces}
The ventral stream task was a same-different unfamiliar face matching forced choice (see figure \ref{fig:faces}), with uniform background, natural background, and occlusions. It is assumed to have no memory demands. In each trial, signatures from the face-tuned cortical component of the architecture (see figure \ref{fig:arch}) are computed for both face images in the test pair.  The model's response was taken to be the thresholded cosine similarity between the two signatures. Strong performance on this task required tolerance to 3D rotation in depth. In the training phase, the model was presented with 320 videos (image sequences), each depicting the rotation of a different individual. The test sets were drawn from the images of the remaining 80 individuals.  The training phase was taken to be a model of visual development. The individuals of the training phase modeled people with which the subject would be highly familiar: parents, friends, etc.

\begin{figure}[tb]
\begin{minipage}{.3\textwidth}
\centering
\includegraphics[height=.2\textheight]{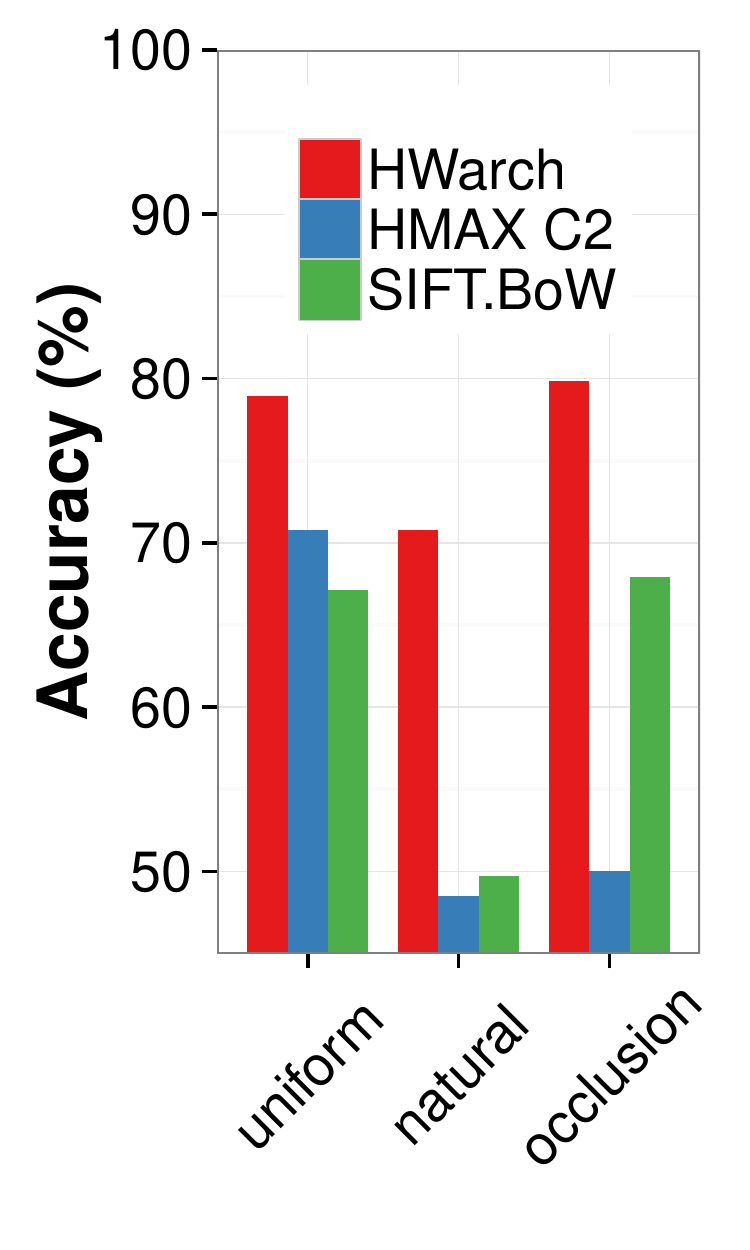}
\captionof{figure}{Ventral stream task results: our model HWarch, outperforms two baselines from \citep{Leibo}. } 
\label{fig:results:ventral}
\end{minipage}\hfill%
\begin{minipage}{.65\textwidth}
\centering
\includegraphics[height=.2\textheight]{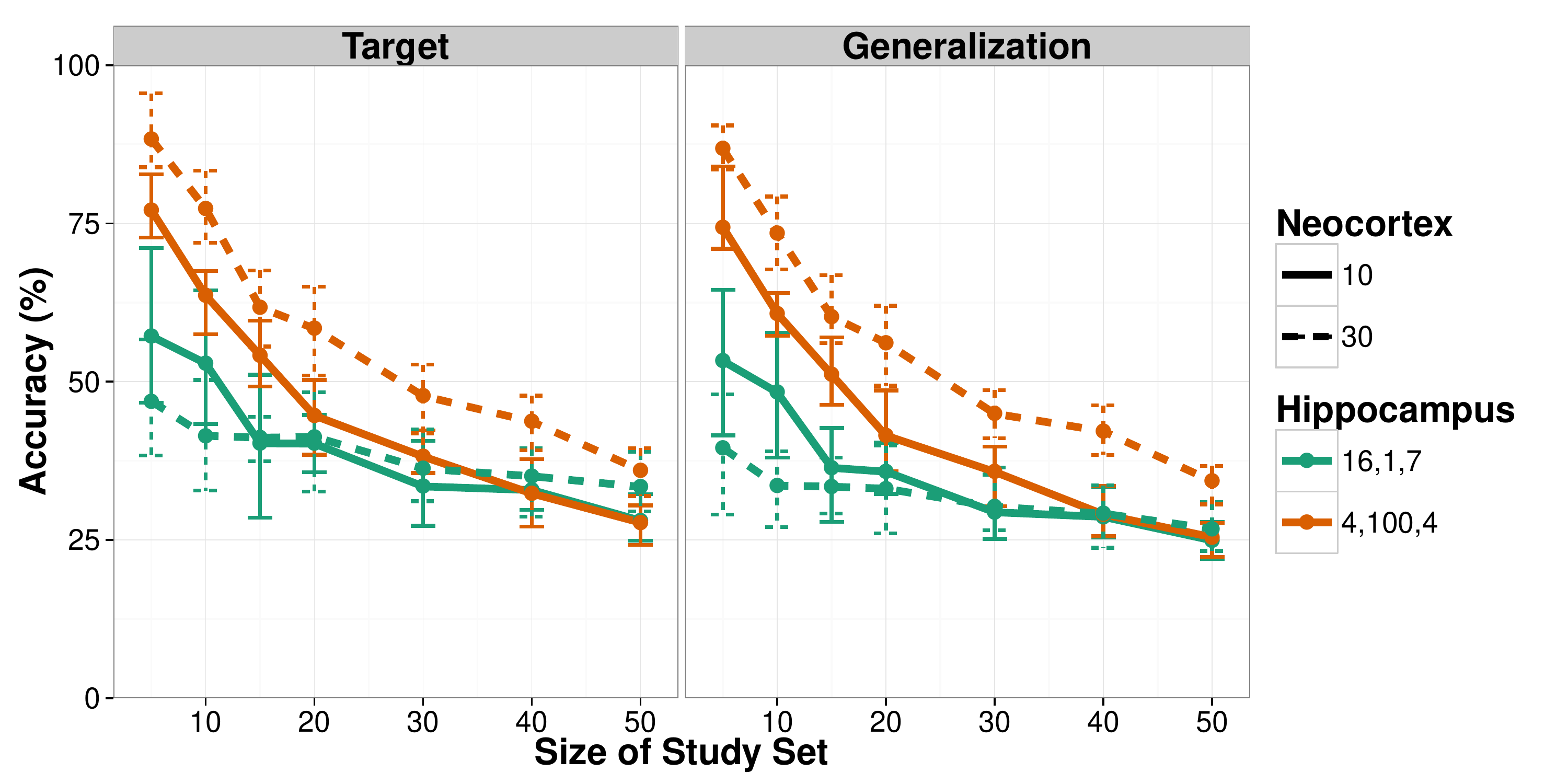} 
\captionof{figure}{MTL task results. Error bars: 25- and 75-percentile over 20 repetitions with different development, study, and testing sets.}
\label{fig:results:MTL}
\end{minipage}
\end{figure}

The interpretation of the training procedure is that high level visual representations are tuned according to a temporal association-based rule. There is evidence from psychophysics \citep{Wallis2001, Cox2005} and neurophysiology \citep{Li2008,Li2010} that ventral stream representations are adapted to temporal correlations in this way. Figure \ref{fig:results:ventral} shows that  our model, HWarch, using the Cortex-1 (FFA-esque) subsystem of figure \ref{fig:arch},  outperforms two baseline feature representations \citep{Leibo} on the yaw rotation invariant same-different matching of unfamiliar faces (SUFR datasets \citep{Leibo}). Note that both baselines used an SVM (supervised training) whereas our model only compared the cosine similarity between each pair of test faces. Under the proposed cortical (PCA) approximation, the templates of S-cells would correspond to projections of the frames onto principal components. It is assumed that this \texttt{INSERT} operation would run slowly and take many interleaved repetitions of the data (though in the case of our experiment we just computed the PCA). The signature computed by a call to \texttt{QUERY} measures the input's similarity to the closest frame of each sequence (assuming $P = \max$). As long as temporally adjacent frames usually depict the same identity face, the signature will remain stable with viewing angle \cite{PartI2013, Liao2013}. The level of accuracy achieved is comparable to similar systems \citep{Leibo2011b, Liao2013, Liao2013a, Leibo2014} that were presented as models of view-tolerant representations in the anterior medial (AM) patch of the macaque face-processing system \citep{freiwald2010functional}.

\subsection{MTL experiment: Recall of face-name associations}
The MTL experiment tested recall of associations between faces and names. Units in the hippocampal layer (fig. \ref{fig:arch}) can be regarded as modeling the multimodal cells described by \citep{quiroga2009explicit}. These cells, discovered in the MTL of human intractable epilepsy patients, respond selectively to visually presented famous faces, e.g., Saddam Hussein's face, the image of his name, and the sound of it spoken (though we do not address the auditory component here). Critically, \citep{quiroga2009explicit} also found that some of these multimodal cells were tuned to the researchers themselves. Since the researchers were unknown to the patients prior to the experiment, the multimodal cells tuned to them must have quickly acquired this selectivity.

This experiment explores multimodal binding. Units in the top layer of figure \ref{fig:arch} can be regarded as modeling the MTL cells recorded by \citep{quiroga2009explicit}. The experiment was divided into three phases. \emph{Development phase:} In first phase, Cortex-1 (the face area) was trained in the exact same way as in experiment 1. Cortex-2 was trained analogously, but in this case, each template book contained a set of template images depicting equivalent words (4-letter names) at a range of positions and fonts. Both cortex-1 and 2 used the PCA approximation. \emph{Study phase: } The training of the hippocampal layer modeled the task of meeting a set of people at an event, say a poster session. That is, each individual consists of a face and a nametag. The two pixel representations are concatenated and encoded in Cortex-1/2. However, the model never sees the nametag or face from all views and the names vary both in retinal location and typeface. Training consisted of \texttt{INSERT}-ing all the images associated with the same individual to an (initially empty) hippocampal HW-module representing the episodic memory of that individual. Since it is assumed that there is not enough time for there to be an effect on cortex, none of the faces or names used to train the cortical layers were used in the study phase. \emph{Test phase:} A test of recalling names from faces, or faces from names. In both cases, the goal is for the maximally activated hippocampal HW-module to be the one containing the representation of the query.

Figure \ref{fig:results:MTL} shows results from a simulation of the episodic recall of names from taxes task, using the full architecture of figure \ref{fig:arch}. The neocortex parameter that we vary is the number of faces used for the cortical model in the first layer. The hippocampal parameters were the WTA hashing parameters: he WTA's metric-determining parameter $K$, the number of hashes $L$ appearing in Alg. \ref{alg:insertquery}, and  the number of $K$-bit integers used per hash $W$, see \citep{Dean2013}.

The assumption underlying this simulation is that hippocampus learns associations between arbitrary items. This simulation tests both the usual case in the memory literature when the probe stimulus (i.e., the query) is exactly one of the items encountered during the study phase, and the less commonly studied case where a correct probe stimulus is a semantically equivalent item. We argue that the latter is the more ecologically valid of the two tasks. There are two mechanisms through which recall decreases with increasing study set size: First, the cortical representation ``compresses'' differences in the input, making them appear more similar. This is useful for generalizing rotation invariance, but problematic for the recall task. Second, the LSH approximation can miss some candidate items when there are too many. The former models ``capacity limits'' due to the cortical afferents while the latter models capacity limits arising from the hippocampus itself.

\section{Discussion}
There are myriad potential gains from thinking about perception and memory in an integrated way. We've already pointed out the additional useful constraints on MTL modeling posed by realistic perceptual inputs, but the benefits of an integrated view are not unidirectional. For example, insight into outstanding questions in vision like the learning of invariant representations may be informed by a modern view of semantic memory, replay and consolidation. The primary role of sensory cortex can be thought of as learning the statistics of the input distribution. However, a learning organism in the real world may not wish to merely be a slave to the statistics of its environment but may want to bias its learning towards inputs that have some relevance to its goals. Hippocampal replay may provide just such a mechanism by preferentially replaying rewarding episodes during sleep and thereby presenting neocortex with more examples from which to learn \citep{singer2009rewarded}. In this way the statistics of the environment can be circumvented.

\newpage
\section{Supplementary Information}

\subsection{Recognition experiment details}
We tested three different datasets from the Subtasks of Unconstrained Face Recognition (SUFR) benchmark collection \citep{Leibo}.  They were 1. yaw rotation with uniform background (shown in Figure 1 of main text)\, 2. yaw rotation with natural image background, and 3. yaw rotation with occlusion (demanding invariance to the presence or absence of sunglasses).  There were 400 individual faces and 19 images of each face (19 different orientations) in task 1 and 2. Task 3 had $38=19\times2$ images per individual. The classifier was the thresholded cosine similarity between the two images to be compared. The threshold was optimized on the training set.

\subsection{Unsupervised versus error-based learning}
We focus on associations that can be learned without explicit error signals. This likely constitutes the majority of the ventral stream's learning \citep{li_neuronal_2012}, but only a part of the MTL's.

\subsection{Feedforward models}
One potential caveat to our proposal is that it neglects generative phenomena and top-down influences.  The framework of this article extends feedforward models of the ventral stream into the medial temporal lobe. Feedforward models may account for neurophysiology results obtained by analyses restricted to a window around the time the first volley of spikes arrives in a region or behavioral effects elicited from brief presentations  ($<$ 100ms) and masking \citep{Thorpe1996, Serre2007, liu2009timing, DiCarlo2012, Isik2014}. They are not expected to capture attentional effects, priming / adaptation effects, mental rotation, and a host of other ventral stream-related phenomena.

When two regions are known to be monosynaptically connected, a feedforward model is the most straightforward hypothesis. Demonstrating a feedforward solution to a difficult computational problem motivates experiments under the aforementioned brief presentation conditions and detailed examination of neuronal response latencies. Note however, it does not constitute a denial of other phenomena.

\subsection{Neuronal selectivity latencies}
In accord with our proposal, selectivity latencies increase from the anterior ventral stream (100-200ms) \citep{Hung2005, liu2009timing} to MTL (200-500ms) \citep{kreiman2000category, mormann2008latency}. It is sometimes claimed that the significantly longer latencies in MTL\footnote{though MTL latencies may be shorter in macaque data \citep{suzuki1997object}} support the hypothesis that ``long loop'' recurrence is necessary to generate selectivity in these regions e.g. \citep{kumaran2012generalization}. Without rejecting that hypothesis, we note that such results are also compatible with a feedforward model. One explanation is that additional integration time per stage is needed to integrate cross-modal inputs, especially if different information processing streams undergo differing numbers of stages prior to MTL. Another possibility is that each MTL area actually contains several processing stages. In this vein, it's notable that MTL areas are typically defined by cytoarchitectural and anatomical criteria rather than physiological criteria as visual areas often are (containing a map of visual space).

\newpage
\section{Acknowledgments}
The authors would like to thank Peter Dayan, Guillaume Desjardins, Greg Wayne, Andrei Rusu, Vlad Mnih, Dharshan Kumaran, and Fernando Pereira for helpful comments on early versions of this manuscript, Ioannis Antonoglou for engineering, and Adam Cain for graphic design.

\bibliographystyle{Science}
\setlength{\bibsep}{0pt plus 0.3ex}
\bibliography{approxHW}

\end{document}

%% file: macros.tex
\usepackage{amsmath, amsthm}
\usepackage{amsfonts}
\usepackage{amssymb}

\DeclareMathOperator*{\argmax}{arg\,max}


%% file: approxHW.bbl
\begin{thebibliography}{10}

\bibitem{bulthoff1992psychophysical}
H.~B\"{u}lthoff, S.~Edelman, {\it Proceedings of the National Academy of
  Sciences\/} {\bf 89}, 60 (1992).

\bibitem{Tarr1995}
M.~J. Tarr, H.~H. B\"{u}lthoff, {\it Journal of Experimental Psychology: Human
  Perception and Performance\/} {\bf 21}, 1494 (1995).

\bibitem{Hung2005}
C.~P. Hung, G.~Kreiman, T.~Poggio, J.~J. DiCarlo, {\it Science\/} {\bf 310},
  863 (2005).

\bibitem{freiwald2010functional}
W.~A. Freiwald, D.~Tsao, {\it Science\/} {\bf 330}, 845 (2010).

\bibitem{Tulving1985}
E.~Tulving, {\it {Elements of episodic memory}\/} (Oxford University Press,
  1985).

\bibitem{Andersen2006}
P.~Andersen, R.~Morris, D.~Amaral, T.~Bliss, J.~O'Keefe, {\it {The hippocampus
  book}\/} (Oxford University Press, 2006).

\bibitem{Hubel1962}
D.~Hubel, T.~Wiesel, {\it The Journal of Physiology\/} {\bf 160}, 106 (1962).

\bibitem{McClelland1995}
J.~L. McClelland, B.~L. McNaughton, R.~C. O'Reilly, {\it Psychological
  review\/} {\bf 102}, 419 (1995).

\bibitem{Norman2003}
K.~A. Norman, R.~C. O'Reilly, {\it Psychological review\/} {\bf 110}, 611
  (2003).

\bibitem{Serre2007}
T.~Serre, A.~Oliva, T.~Poggio, {\it Proceedings of the National Academy of
  Sciences of the United States of America\/} {\bf 104}, 6424 (2007).

\bibitem{Suzuki1994a}
W.~A. Suzuki, D.~G. Amaral, {\it Journal of Comparative Neurology\/} {\bf 350},
  497 (1994).

\bibitem{Kali2004}
S.~Kali, P.~Dayan, {\it Nature Neuroscience\/} {\bf 7}, 286 (2004).

\bibitem{Riesenhuber1999}
M.~Riesenhuber, T.~Poggio, {\it Nature Neuroscience\/} {\bf 2}, 1019 (1999).

\bibitem{Lee2003}
T.~S. Lee, D.~B. Mumford, {\it Journal of the Optical Society of America\/}
  {\bf 20}, 1434 (2003).

\bibitem{Rolls2012}
E.~Rolls, {\it Frontiers in Computational Neuroscience\/} {\bf 6} (2012).

\bibitem{DiCarlo2012}
J.~J. DiCarlo, D.~Zoccolan, N.~C. Rust, {\it Neuron\/} {\bf 73}, 415 (2012).

\bibitem{sadato1996activation}
N.~Sadato, {\it et~al.\/}, {\it Nature\/} {\bf 380}, 526 (1996).

\bibitem{roder2002speech}
B.~R\"{o}der, O.~Stock, S.~Bien, H.~Neville, F.~R\"{o}sler, {\it European
  Journal of Neuroscience\/} {\bf 16}, 930 (2002).

\bibitem{sur1988experimentally}
M.~Sur, P.~Garraghty, A.~Roe, {\it Science\/} {\bf 242}, 1437 (1988).

\bibitem{LeCun1989}
Y.~LeCun, {\it et~al.\/}, {\it Neural computation\/} {\bf 1}, 541 (1989).

\bibitem{Fukushima1980}
K.~Fukushima, {\it Biological Cybernetics\/} {\bf 36}, 193 (1980).

\bibitem{Krizhevsky2012}
A.~Krizhevsky, I.~Sutskever, G.~Hinton, {\it Advances in neural information
  processing systems\/} (Lake Tahoe, CA, 2012), vol.~25, pp. 1106--1114.

\bibitem{Abdel-Hamid2012}
O.~Abdel-Hamid, A.~Mohamed, H.~Jiang, G.~Penn, {\it IEEE International
  Conference on Acoustics, Speech and Signal Processing (ICASSP)\/} (2012), pp.
  4277--4280.

\bibitem{marr2010vision}
D.~Marr, {\it {Vision: A computational investigation into the human
  representation and processing of visual information}\/} (Henry Holt and Co.,
  Inc., New York, NY, 1982).

\bibitem{PartI2013}
F.~Anselmi, {\it et~al.\/}, {\it arXiv:1311.4158v3 [cs.CV]\/}  (2013).

\bibitem{lecun1995convolutional}
Y.~LeCun, Y.~Bengio, {\it The handbook of brain theory and neural networks\/}
  pp. 255--258 (1995).

\bibitem{Serre2007a}
T.~Serre, L.~Wolf, S.~Bileschi, M.~Riesenhuber, T.~Poggio, {\it IEEE
  Transactions on Pattern Analysis and Machine Intelligence\/} {\bf 29}, 411
  (2007).

\bibitem{Fix1951}
E.~Fix, J.~L. Hodges, {\it Defense Technical Information Center (DTIC)
  report\/} {\bf ADA800276} (1951).

\bibitem{coates2012emergence}
A.~Coates, A.~Karpathy, A.~Y. Ng, {\it Advances in Neural Information
  Processing Systems\/} (2012), pp. 2681--2689.

\bibitem{Isik2012}
L.~Isik, J.~Z. Leibo, T.~Poggio, {\it Front. Comput. Neurosci.\/} {\bf 6}
  (2012).

\bibitem{liao2014unsupervised}
Q.~Liao, J.~Z. Leibo, T.~Poggio, {\it arXiv preprint arxiv:1409.3879\/}
  (2014).

\bibitem{Foldiak1991}
P.~F\"{o}ldi\'{a}k, {\it Neural Computation\/} {\bf 3}, 194 (1991).

\bibitem{Liao2013}
Q.~Liao, J.~Z. Leibo, T.~Poggio, {\it Advances in Neural Information Processing
  Systems (NIPS)\/} (Lake Tahoe, CA, 2013).

\bibitem{eichenbaum1993memory}
H.~Eichenbaum, {\it Memory, amnesia, and the hippocampal system\/} (MIT press,
  1993).

\bibitem{quiroga2009explicit}
R.~Q. Quiroga, A.~Kraskov, C.~Koch, I.~Fried, {\it Current Biology\/} {\bf 19},
  1308 (2009).

\bibitem{riesenhuber2002neural}
M.~Riesenhuber, T.~Poggio, {\it Current Opinion in Neurobiology\/} {\bf 12},
  162 (2002).

\bibitem{oja1982simplified}
E.~Oja, {\it Journal of mathematical biology\/} {\bf 15}, 267 (1982).

\bibitem{sanger1989optimal}
T.~Sanger, {\it Neural networks\/} {\bf 2}, 459 (1989).

\bibitem{oja1992principal}
E.~Oja, {\it Neural Networks\/} {\bf 5}, 927 (1992).

\bibitem{Johnson1984}
W.~B. Johnson, J.~Lindenstrauss, {\it Contemporary mathematics\/} {\bf 26}, 189
  (1984).

\bibitem{wiskott2002slow}
L.~Wiskott, T.~Sejnowski, {\it Neural computation\/} {\bf 14}, 715 (2002).

\bibitem{Dean2013}
T.~Dean, {\it et~al.\/}, {\it IEEE Conference on Computer Vision and Pattern
  Recognition (CVPR)\/} (IEEE, Portland, OR, USA, 2013), pp. 1814--1821.

\bibitem{Yagnik2011}
J.~Yagnik, D.~Strelow, D.~A. Ross, R.-s. Lin, {\it IEEE International
  Conference on Computer Vision (ICCV)\/} (IEEE, Barcelona, Spain, 2011), pp.
  2431--2438.

\bibitem{Andoni2006}
A.~Andoni, P.~Indyk, {\it 47th Annual IEEE Symposium on Foundations of Computer
  Science\/} (2006), pp. 459--468.

\bibitem{Wallis2001}
G.~Wallis, H.~H. B\"{u}lthoff, {\it Proceedings of the National Academy of
  Sciences of the United States of America\/} {\bf 98}, 4800 (2001).

\bibitem{Cox2005}
D.~Cox, P.~Meier, N.~Oertelt, J.~J. DiCarlo, {\it Nature Neuroscience\/} {\bf
  8}, 1145 (2005).

\bibitem{Li2008}
N.~Li, J.~J. DiCarlo, {\it Science\/} {\bf 321}, 1502 (2008).

\bibitem{Li2010}
N.~Li, J.~J. DiCarlo, {\it Neuron\/} {\bf 67}, 1062 (2010).

\bibitem{Leibo}
J.~Z. Leibo, Q.~Liao, T.~Poggio, {\it International Joint Conference on
  Computer Vision, Imaging and Computer Graphics, VISAPP\/} (SCITEPRESS,
  Lisbon, Portugal, 2014).

\bibitem{Leibo2011b}
J.~Z. Leibo, J.~Mutch, T.~Poggio, {\it Advances in Neural Information
  Processing Systems (NIPS)\/} (Granada, Spain, 2011).

\bibitem{Liao2013a}
Q.~Liao, J.~Z. Leibo, Y.~Mroueh, T.~Poggio, {\it arXiv preprint
  arXiv:1311.4082\/}  (2014).

\bibitem{Leibo2014}
J.~Z. Leibo, Q.~Liao, F.~Anselmi, T.~Poggio, {The invariance hypothesis implies
  domain-specific regions in visual cortex}, {\it Tech. rep.\/}, CBMM (2014).

\bibitem{singer2009rewarded}
A.~C. Singer, L.~M. Frank, {\it Neuron\/} {\bf 64}, 910 (2009).

\bibitem{li_neuronal_2012}
N.~Li, J.~J. {DiCarlo}, {\it The Journal of Neuroscience\/} {\bf 32},
  6611–6620 (2012).

\bibitem{Thorpe1996}
S.~Thorpe, D.~Fize, C.~Marlot, {\it Nature\/} {\bf 381}, 520 (1996).

\bibitem{liu2009timing}
H.~Liu, Y.~Agam, J.~R. Madsen, G.~Kreiman, {\it Neuron\/} {\bf 62}, 281 (2009).

\bibitem{Isik2014}
L.~Isik, E.~M. Meyers, J.~Z. Leibo, T.~Poggio, {\it Journal of
  Neurophysiology\/} {\bf 111}, 91 (2014).

\bibitem{kreiman2000category}
G.~Kreiman, C.~Koch, I.~Fried, {\it Nature neuroscience\/} {\bf 3}, 946 (2000).

\bibitem{mormann2008latency}
F.~Mormann, {\it et~al.\/}, {\it The Journal of Neuroscience\/} {\bf 28}, 8865
  (2008).

\bibitem{suzuki1997object}
W.~A. Suzuki, E.~K. Miller, R.~Desimone, {\it Journal of neurophysiology\/}
  {\bf 78}, 1062 (1997).

\bibitem{kumaran2012generalization}
D.~Kumaran, J.~L. McClelland, {\it Psychological review\/} {\bf 119}, 573
  (2012).

\end{thebibliography}
